\pdfoutput=1
\documentclass[letterpaper, 10 pt, conference]{ieeeconf}  
\usepackage{graphicx}
\usepackage{indentfirst}
\usepackage{amsfonts,amssymb}
\usepackage{svg}
\usepackage{float}
\usepackage{multirow}
\usepackage{bm}
\usepackage{subfigure}
\usepackage{url}
\usepackage{hyperref}
\usepackage{cite}
\usepackage{amsmath}
\usepackage{color} 
\usepackage{booktabs}

\IEEEoverridecommandlockouts                              

\title{\LARGE \bf
Autonomous Exploration and Mapping for Mobile Robots via Cumulative Curriculum Reinforcement Learning
}

\author{Zhi Li$^{1}$, Jinghao Xin$^{1}$, and Ning Li$^{1}$
\thanks{*This work is supported by National Nature Science Foundation under Grant (62273230).}
\thanks{Zhi Li, Jinghao Xin and Ning Li are with Department of Automation, Shanghai Jiao Tong University, Shanghai 200240, P.R. China, and also with  Key Laboratory of System Control and Information Processing, Ministry of Education of China, Shanghai 200240, China, and also with Shanghai Engineering Research Center of Intelligent Control and Management, Shanghai 200240, China (E-mail: lizhibeaman, xjhzsj2019, ning\_li@sjtu.edu.cn)}%
}

\begin{document}

\maketitle
\thispagestyle{empty}
\pagestyle{empty}

\begin{abstract}
Deep reinforcement learning (DRL) has been widely applied in autonomous exploration and mapping tasks, but often struggles with the challenges of sampling efficiency,  poor adaptability to unknown map sizes, and slow simulation speed.
To speed up convergence, we combine curriculum learning (CL) with DRL, and first propose a Cumulative Curriculum Reinforcement Learning (CCRL) training framework to alleviate the issue of catastrophic forgetting faced by general CL.
Besides, we present a novel state representation, which considers a local egocentric map and a global exploration map resized to the fixed dimension, so as to flexibly adapt to environments with various sizes and shapes. 
Additionally, for facilitating the fast training of DRL models, we develop a lightweight grid-based simulator, which can substantially accelerate simulation compared to popular robot simulation platforms such as Gazebo. Based on the customized simulator, comprehensive experiments have been conducted, and the results show that the CCRL framework not only mitigates the catastrophic forgetting problem, but also improves the sample efficiency and generalization of DRL models, compared to general CL as well as without a curriculum.
Our code is available at \url{https://github.com/BeamanLi/CCRL_Exploration}.
\end{abstract}

\section{Introduction}

Autonomous exploration and mapping means that mobile robots actively explore the \emph{priori} unknown environment without collisions while constructing a map of the surroundings as entirely as possible\cite{lizhi2022-RCAR}, which has been widely applied to military reconnaissance\cite{garaffa2021-RL-exploration-survey}, search and rescue\cite{niroui2019-search-rescue}, planetary exploration \cite{koutras2021-planetary-exploration}, and other fields.

Traditional autonomous exploration methods mainly include frontier-based \cite{yamauchi1997frontier} and information-based\cite{information-based} strategies. The former determines the robot's moving target according to frontiers, which are defined as the boundary regions between the free and unknown space. The latter constructs complicated optimization problems based on mutual information.  In general, computational complexity and reliance on handcrafted expert features limit the applications of traditional exploration methods in the real world\cite{garaffa2021-RL-exploration-survey}.

Thanks to breakthroughs in deep reinforcement learning (DRL)\cite{mnih2015DQN,silver2016AlphaGo} in the last decade, some researchers have applied DRL to autonomous exploration tasks. Most previous DRL-based exploration models\cite{niroui2019-search-rescue,zhu2018-DRL-exploration-office,chen2019-DRL-exploration-self-learning,wang2021-DRL-exploration-SpatialActionMaps} need to be combined with traditional exploration or navigation algorithms, referred to as \emph{2-stage} strategies \cite{garaffa2021-RL-exploration-survey}, hence still suffer from the  model complexity issue mentioned above.
In our previous work\cite{lizhi2022-RCAR}, we proposed an \emph{end-to-end} DRL-based exploration model that directly outputs discrete control commands, but with a major limitation of poor adaptability to diverse environment sizes. 
In this paper, we introduce an improved map representation that considers a local egocentric map and a global exploration map resized to the fixed dimension, allowing flexible adaptation to maps with varied sizes and shapes.

One of the critical challenges when applying DRL to robotic systems,  especially in the end-to-end training paradigm, is \emph{sample efficiency}. Due to complicated dynamics and high-dimensional image input, training a reinforcement learning (RL) agent to learn an optimal policy may require millions of  interactions with the environment\cite{narvekar2020-CL-survey}. This issue can be aggravated by the slow simulation speed when using popular robot simulation platforms such as Gazebo\cite{koenig2004-Gazebo}. The enormous sample steps and slow simulation speed make the training time of  DRL models in robotic applications prohibitive. To address the above issues, we propose the following solutions:

On the one hand, we apply curriculum learning (CL) \cite{bengio2009curriculum} to improve sample efficiency and speed up convergence, which starts learning from simple tasks and gradually increases the difficulty of the tasks\cite{narvekar2020-CL-survey}. However, when combining CL with RL, especially in the context of modern DRL, a crucial dilemma is \emph{catastrophic forgetting}\cite{rusu2016progressive}: the knowledge learned from previous tasks may be gradually lost when training on new tasks. In order to alleviate this issue, we first propose a \textbf{C}umulative \textbf{C}urriculum \textbf{R}einforcement \textbf{L}earning \textbf{(CCRL)} training framework: Instead of being directly transferred to the following more difficult environment when performing a curriculum task switch,  the agent will interact with the vectorized environments composed of the previous and new tasks with the aid of  \emph{AsyncVectorEnv} in OpenAI Gym\cite{Gym}.

On the other hand, we design a lightweight grid-based autonomous exploration simulator specifically for end-to-end training and CL, which contains a series of training maps with progressively increasing difficulty based on four typical map features adapted from \cite{xu2022-explore-bench}. Our grid-based simulator supports customized maps and can significantly  accelerate the simulation compared to Gazebo.

The main contributions of this paper are summarized here:

(1) We propose a Cumulative Curriculum Reinforcement Learning (CCRL) training framework to moderate the catastrophic forgetting issue faced by general CL while improving the sample efficiency and generalization of DRL models.

(2) We present an end-to-end DRL-based autonomous exploration and mapping model with a size-adaptive map representation, which can flexibly adapt to environments with different sizes and shapes.

(3) We customize a concise grid-based autonomous exploration simulator specifically for end-to-end training and curriculum learning, facilitating fast implementation, verification, and comparison of DRL algorithms.

\section{Related Work}
\textbf{Traditional Exploration Methods.}
The frontier-based exploration is most widespread among the traditional exploration methods, first proposed by Yamauchi et al.\cite{yamauchi1997frontier}, where the robot always naively navigated to the nearest frontier. In the following decades, various improvements to the frontier-based strategy have been developed, mainly focusing on how to select the most promising frontier,  including path cost\cite{mei2006-cost-path}, information gain\cite{bai2016-information-gain}, potential field\cite{yu2021-potential-field}, \emph{etc}. Traditional exploration methods often rely on handcrafted expert features and strong assumptions about specific tasks, which decreases the adaptive capacity for diverse unknown environments. Besides, as the map size and robot action space expand, the computational complexity and decision time of traditional methods will grow substantially\cite{chen2020-graph}. 

\textbf{DRL-based Exploration Methods.}
Niroui et al.\cite{niroui2019-search-rescue} combined DRL with the traditional frontier-based methods, where the A3C\cite{mnih2016A3C} model output the weight parameters of  each frontier. The frontier with the lowest cost calculated by the predefined cost function would be assigned to the robot. In \cite{zhu2018-DRL-exploration-office}, an A3C policy selected one of the six sector subregions centered on the robot as the next visiting direction, and the robot navigated to the target in this candidate subregion determined by the next-best-view algorithm \cite{NBV}. Wang et al. \cite{wang2021-DRL-exploration-SpatialActionMaps} presented an autonomous exploration method based on spatial action maps, where action commands could be represented as pixels on the map. The DDQN\cite{van2016DDQN} algorithm was employed to encode the Q-value of each pixel, and the robot chose to move to the target point with the highest Q-value. 
Since the above DRL-based models need to be combined with traditional exploration or navigation algorithms, the problem of computational burden described above still exists.
In addition, most previous DRL models use global or local grid maps with fixed dimensions as the state space, which are less adaptable to varying environment sizes. In this paper, we propose a size-adaptive end-to-end DRL-based exploration model, which directly outputs discrete control commands and flexibly adapts to diverse maps.

\textbf{Curriculum Learning.}
To speed up convergence and improve training performance or sample efficiency, curriculum learning techniques have been widely used in DRL-based mobile robot navigation (e.g., gradually increasing the number of obstacles and the distance between the robot and the target\cite{chen2020-curr-nav}) and exploration (e.g., training an agent to explore environments with gradually growing sizes\cite{zhang2017-curr-explore-1,chen2019-curr-explore-2})  tasks. To address the catastrophic forgetting problem existing in the general CL, Rusu et al.\cite{rusu2016progressive} proposed a ``progressive neural network'', which trained a new network ``column'' for each new task. When training subsequent columns, parameters from previous columns would be frozen. The main limitation of this method is that the number of model parameters and the inference time will increase with the number of tasks. In contrast, our CCRL framework only increases the number of vectorized environments during training, and the neural network structure is consistently fixed.

\section{Methods}

\subsection{End-to-end DRL-based Autonomous Exploration and Mapping}

\subsubsection{Problem Formulation}
We formulate the autonomous exploration and mapping task as a Markov decision process (MDP). At timestep $t$, the agent observes the state of the environment $s_t \in \mathcal{S}$, takes the action $a_t \in \mathcal{A}$ according to the policy $\pi$, receives the reward $r_t$, and then transits to the next state  $s_{t+1} \in \mathcal{S}$, where $\mathcal{S}$ is the state space, $\mathcal{A}$ is the action space. The goal of the RL agent is to learn an optimal policy to maximize the expectation of discounted cumulative rewards $\mathbb{E}_{\pi}[\sum\limits_{t =0}^{+\infty} \gamma ^ t r_{t}]$, where $\gamma \in [0,1]$ is the discount factor. 
In this paper, we implement Proximal Policy Optimization (PPO)\cite{schulman2017-PPO} as the underlying DRL algorithm, which is a popular and powerful on-policy DRL algorithm and has been widely applied in locomotion control\cite{DPPO}, video games\cite{raileanu2020-DrAC}, robot navigation \cite{PPO-navigation}, \emph{etc}. The overview of our end-to-end DRL-based autonomous exploration and mapping model is shown in Fig. 1. 

\begin{figure}[t]
\setlength{\abovecaptionskip}{0cm}
\center
\includegraphics[width=0.49\textwidth]{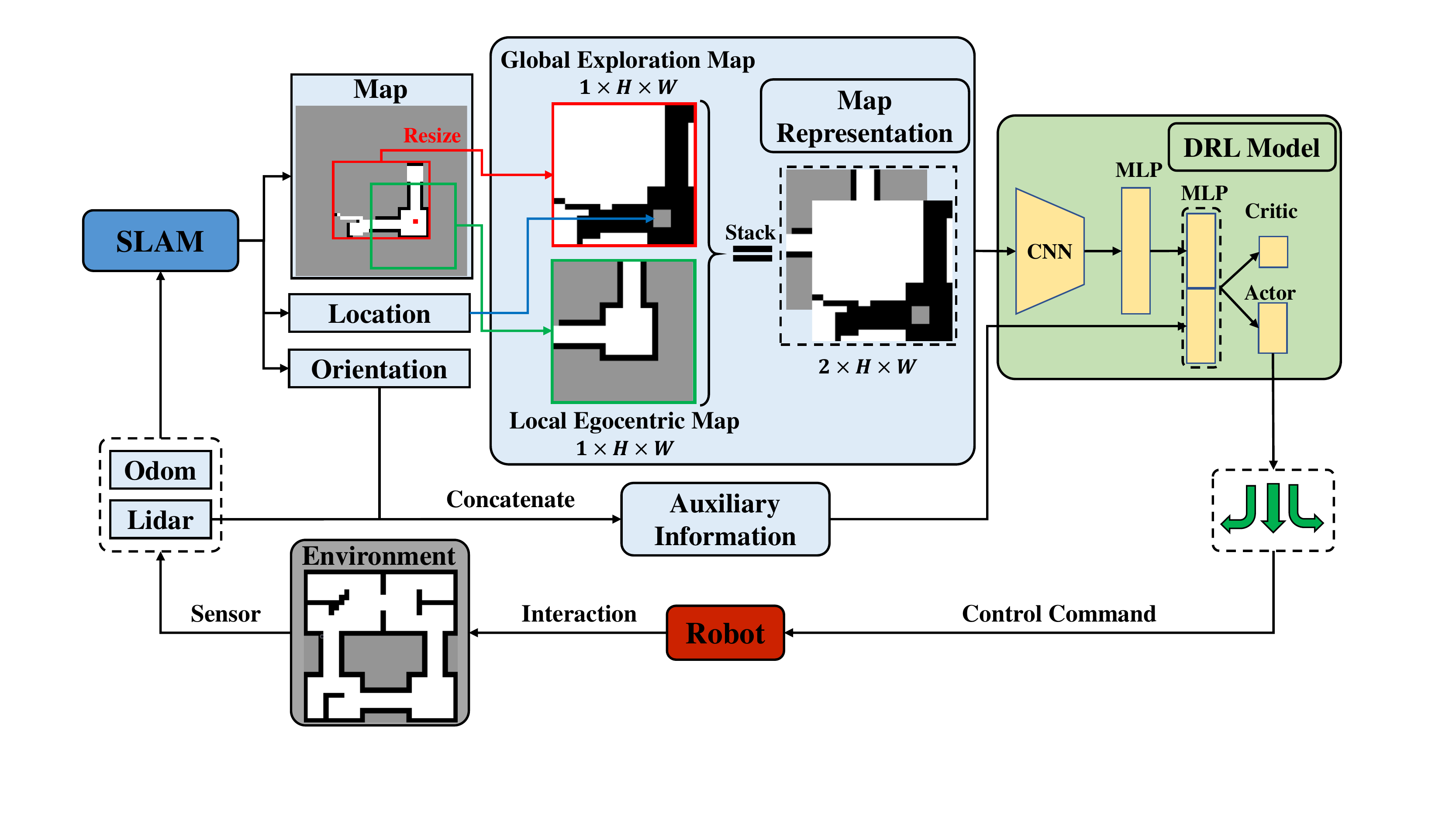}
\caption{\textbf{Overview of our end-to-end DRL-based autonomous exploration and mapping model.}}
\label{fig1}
\vspace{-0.5cm}
\end{figure}

\subsubsection{State Space}
We propose a novel map representation that combines local and global map information, as well as the agent's location, which can adapt to different sizes of environments.

\textbf{Local Egocentric Map (LEM)}.
At timestep $t$, we employ SLAM (Simultaneous Localization and Mapping) module to construct the 2D occupied grid map $M_t$ and estimate the agent's location $(x_t, y_t)$ and orientation $\theta_t$. A local egocentric map (LEM) $M^{l}_t$ can be extracted from $M_t$ with the fixed dimension of $H \times W$, where the pixel values of the occupied, unknown and free grids are 255, 128, and 0, respectively. The LEM contains only local information within a limited field of view centered on the agent.

\textbf{Global Exploration Map (GEM)}.
We extract the maximum rectangular boundary $M^{b}_t$ of the explored region from $M_t$, where all occupied and free grids are termed as the \emph{explored} state, represented by 255, and unknown grids are termed as the \emph{unexplored} state, represented by 0. In our DRL model, the Convolutional Neural Network (CNN) is used as an encoder to extract features from the map representation. Standard CNN can only process image inputs with fixed dimensions, while the size and shape of the environment is \emph{priori} unknown, and the dimension of  $M^{b}_t$ may keep changing during the exploration. Therefore, we use the nearest-neighbor interpolation to resize $M^{b}_t$ to the same dimension ($H \times W$) as LEM, and mark the relative location of the agent on it with the pixel value of 128 and the dimension of $D \times D$. We refer to this scaled map as the global exploration map (GEM), denoted as $M^{g}_t$, which provides global perceptual information and ensures the dimensions of the images fed into the CNN are constant. At last, we stack $M^{l}_t$ and $M^{g}_t$ in the dimension of the channels, resulting in the final map representation with the dimension of  $2  \times H \times W$.

\textbf{Auxiliary Information}.
In addition to the map representation described above, the state space also includes a vector consisting of the lidar ranging results and the agent's orientation $\theta_t$, where the former is useful to assist the agent's obstacle avoidance, and the latter is necessary for the agent to perceive its own direction in the environment.

\subsubsection{Action Space}
Due to the end-to-end training paradigm of our DRL model, the action space comprises three discrete control commands: straight forward, turn left, and turn right.

\subsubsection{Reward Function}
The reward function is shown in equation (1):
\begin{equation}
r_t=r_t^e+r_t^s+r_t^c,
\end{equation}
which contains the following three components:

\textbf{Encouraging exploration}. Let $\rho_t$ be the map exploration rate at timestep $t$. If the agent explores the new region, the reward is proportional to $\rho_t^2-\rho_{t-1}^2$, which gives a larger reward in the later stage of exploration. 
Meanwhile, to prevent the negative impact of excessive reward on the training of DRL models, the reward will be clipped into $[0,1]$. Otherwise, the agent will receive a minor penalty.
\begin{equation}
r_t^e= 
\begin{cases}
\operatorname{clip}\left(\left(\rho_t^2-\rho_{t-1}^2\right) \times 10,0,1\right) & \text {if } \rho_t>\rho_{t-1}, \\
 -0.005 & \text {otherwise.}
\end{cases}
\end{equation}

\textbf{Successful exploration}. If $\rho_t  \geq 0.99$, we can consider that the exploration task has been accomplished, the agent will receive a bonus of $+1$, and the episode will be terminated.
\begin{equation}
r_t^s=
\begin{cases}1 & \text {if } \rho_t  \geq  0.99, \\ 
0 & \text {otherwise.}
\end{cases}
\end{equation}

\textbf{Obstacle avoidance}. If the agent collides with obstacles or walls, it will receive a penalty of $-1$, and the episode will be terminated.
\begin{equation}
r_t^c=
\begin{cases}
-1 & \text {if collision}, \\ 
0 & \text {otherwise.}
\end{cases}
\end{equation}

\subsection{Grid-based Autonomous Exploration Simulator}
Popular robot simulation platforms, such as Gazebo, are capable of simulating realistic physical properties, but the extremely slow simulation speed prohibits the fast training and evaluation of DRL algorithms.
A grid-based autonomous exploration simulator is proposed in \cite{xu2022-explore-bench}, but it can only be used to train 2-stage DRL-based exploration models\cite{niroui2019-search-rescue,zhu2018-DRL-exploration-office,chen2019-DRL-exploration-self-learning,wang2021-DRL-exploration-SpatialActionMaps}. Inspired by this, we design a more concise and lightweight grid-based autonomous exploration simulator customized for end-to-end training, where the robot is abstracted as a pixel in the grid world. We use the Ray-tracing algorithm\cite{Ray-tracing} to simulate the scanning and mapping process of 2D-lidar, and provide the ground-truth location and orientation of the agent directly, simplifying the slow and computationally complex SLAM process in Gazebo. Besides, the agent's movements in the grid world (straight forward for one grid, turn left $90^\circ$, turn right $90^\circ$) can be considered to be completed instantaneously, replacing the time-consuming moving process in Gazebo. By testing, the single-step simulation time in our proposed grid-based autonomous exploration simulator is about 0.0025s, which is much faster than Gazebo.

Furthermore, our grid-based simulator supports diverse customized maps. In this paper, we design a set of progressively more difficult training maps, as shown in Fig. 2(a), based on four typical map features adapted from \cite{xu2022-explore-bench}, which establish the foundation for the following curriculum learning. In addition, we also build a set of test maps with different sizes and layouts to evaluate the generalization of the DRL models, as shown in Fig. 2(b).
\begin{figure}[t]
\setlength{\abovecaptionskip}{0cm}
\center
\includegraphics[width=0.475\textwidth]{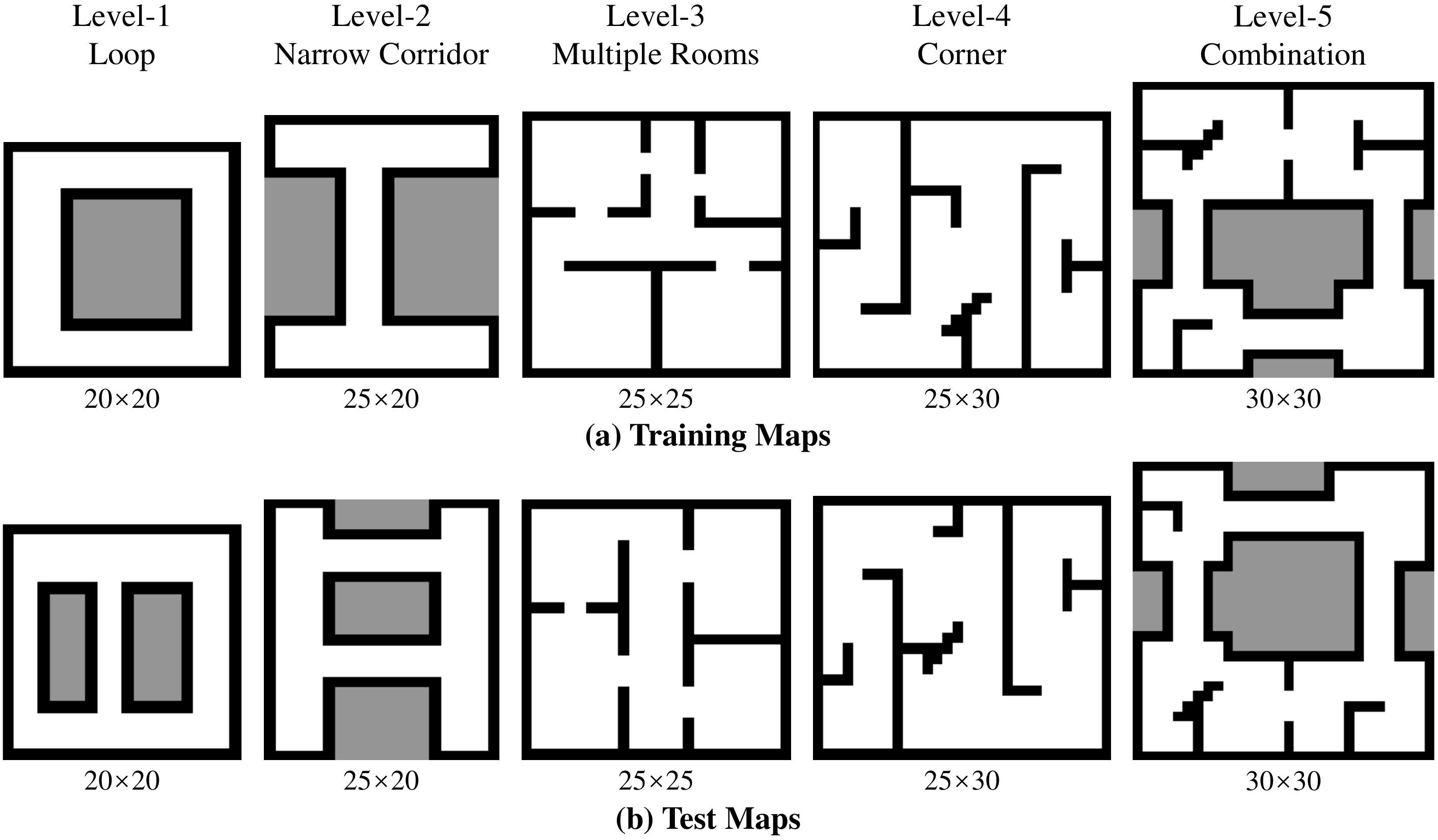}
\caption{\textbf{Grid-based autonomous exploration simulator}: (a) is a set of progressively more difficult training maps for curriculum learning, including four typical map features: \emph{loop} (level-1),  \emph{narrow corridor} (level-2),  \emph{multiple rooms} (level-3),  \emph{corner} (level-4),  and their combination: \emph{combination} (level-5); (b) is a set of test maps with different sizes and layouts. In all maps, black, gray and white grids represent obstacles (or walls), unknown and free areas, with pixel values of 255, 128 and 0, respectively.}
\label{fig2}
\vspace{-0.5cm}
\end{figure}

\subsection{Cumulative Curriculum Reinforcement Learning}
To speed up convergence and improve sample efficiency, we apply curriculum learning to the training of DRL models based on the simulation environments in Fig. 2.  A prevalent problem when combining CL with DRL is catastrophic forgetting\cite{rusu2016progressive}. The reason is that the weights of the neural network optimized for the previous tasks have to be partially modified so as to meet the optimization objectives of the new tasks, which usually results in a deteriorated performance on the original tasks\cite{narvekar2020-CL-survey}. In order to mitigate this issue, we first propose a \textbf{C}umulative \textbf{C}urriculum \textbf{R}einforcement \textbf{L}earning \textbf{(CCRL)} training framework, as shown in \textbf{Algorithm 1}. 

\begin{table}[H]\scriptsize
\vspace{-0.1cm}
\setlength\tabcolsep{1pt}
\label{tab1}
\centering
\begin{tabular}{r l}
\toprule[1pt]
\multicolumn{2}{l}{\textbf{Algorithm 1} Cumulative Curriculum Reinforcement Learning (CCRL)} \\
\midrule[0.5pt]
1:&Initialize the DRL model $\mathcal{M}$\\
2:&Let $\left\{E_1, ..., E_K\right\}$ be a sequence of progressively more difficult environments\\
3:&Vectorize each environment to get N parallel copies $\left\{E_1^N, ..., E_K^N \right\}$\\
&(\emph{This step is not necessary, here $N \geq 1$. In practice, we find that vectorizing }\\
&\emph{each map can make full use of computing  resources and speed up sampling.})\\
4:&Initialize the vectorized enviromments $E=\varnothing$ \\
5:&\textbf{for} $i=1, 2, ..., K$  \textbf{do}\\
6:&\quad $E=E \cup E_i^N$\\
& \quad  (\emph{$\cup$ means integrating previous and new tasks into the vectorized environments})\\
7:&\quad \textbf{while not} A predefined performance criterion is satisfied \textbf{do}\\
8:&\qquad Collect transitions on $E$ and optimize the DRL model $\mathcal{M}$\\
9:&\quad \textbf{end while}\\
10:&\textbf{end for} \\
\bottomrule[1pt]
\vspace{-0.5cm}
\end{tabular}
\end{table}

The main difference between our CCRL and general CL is the concept of  ``\textbf{cumulative}'': When the training process switches from the former stage to the next, instead of being directly transferred to the following more complex environment, the agent will interact with vectorized environments composed of both historical and new tasks with the aid of  \emph{AsyncVectorEnv} in OpenAI Gym\cite{Gym}. 
Vectorized environments run multiple independent copies of the same environment in parallel, take a batch of actions as input, and return a batch of observations and rewards,  which is particularly efficient when using neural networks to process batch data. The DRL model will be optimized based on the transitions collected from past and new environments, enabling the agent to learn additional skills on the new task without forgetting the knowledge acquired from the past.

\begin{figure}[t]
\setlength{\abovecaptionskip}{0cm}
\center
\includegraphics[width=0.475\textwidth]{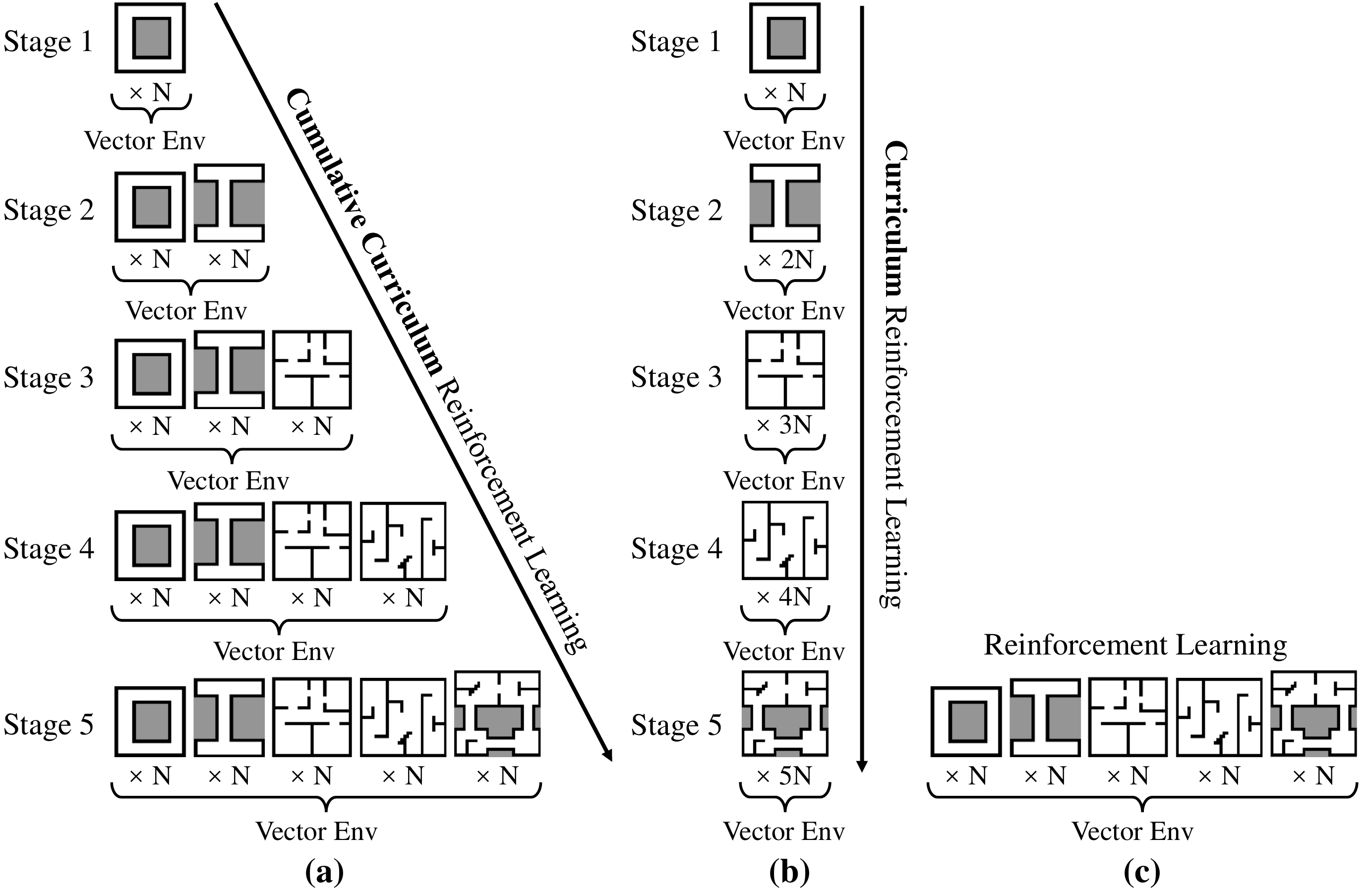}
\caption{(a) is our proposed Cumulative Curriculum Reinforcement Learning (CCRL). When performing a curriculum task switch, the previous and new environments will be integrated into the vectorized environments; (b) is the classical CL paradigm; (c) is the general RL without a curriculum.}
\label{fig3}
\vspace{-0.5cm}
\end{figure}

An essential advantage of our proposed CCRL framework is that it can be easily integrated with mainstream DRL algorithms. In this paper, we combine the PPO algorithm with the CCRL framework, named \textbf{CCPPO} (Cumulative Curriculum PPO), as shown in Fig. 3(a). As a comparison, we also train PPO with general CL (Fig. 3(b)) and without a curriculum (Fig. 3(c)),  named \textbf{CPPO} (Curriculum PPO) and \textbf{PPO}, respectively. It is worth noting that the number of vectorized environments per stage in CPPO is the same as in CCPPO for a fair comparison.

\section{Experiment}
In this section, we train CCPPO, CPPO and PPO algorithms in our grid-based simulator, taking full advantage of its rapidity for fast implementation, evaluation and comparison of  DRL algorithms.

\subsection{Basic Experimental Settings}
To be fair, all DRL algorithms in the training process use the same hyperparameters, which can be found in our open-source code. Moreover, the following experimental settings are common to all algorithms and environments.
\subsubsection{Improving generalization}
To improve the generalization of DRL algorithms, we adopt the following three tricks:
\begin{itemize}
\item Before the beginning of each episode, the location and orientation of the agent will be randomly initialized.
\item Before the beginning of each episode, four obstacles will be randomly placed in the environment.
\item The data augmentation is implemented by rotating the map representation in the state space, as in \cite{raileanu2020-DrAC}.
\end{itemize}

\subsubsection{State space settings}
The dimensions of the map representation and agent's location are set to $H=W=24$ and $D=3$. The scanning angle range of the lidar is $270^\circ$ with a resolution of $9^\circ$. The map representation and auxiliary information are both normalized into $[0,1]$. The number of vectorized environments in each map is set to $N=4$.

\subsubsection{Criterion for switching to the next training stage}
In the curriculum learning, when the average of the map exploration rate of the last 10 evaluations on the current training map (for CCPPO, on the current highest level map) exceeds 0.95, it will switch to the next training stage.

\subsection{Training Stage}
\subsubsection{CCPPO vs CPPO}
The training curves for CCPPO and CPPO are shown in Fig. 4(a). It can be found that when CPPO switches to the following more difficult map, the map exploration rates on the previous levels will gradually decrease, especially on more challenging levels (such as level-3 and level-4), so-called catastrophic forgetting. As a contrast, in CCPPO, the map exploration rates on all levels can converge to nearly 1.0 simultaneously, effectively alleviating the problem of catastrophic forgetting.

In addition, in the last three levels, the initial map exploration rates (indicated by the horizontal dashed line in the figure) of CCPPO are all higher than CPPO. This demonstrates that CCRL can ``accumulate'' knowledge learned from previous tasks and transfer them among different levels, so as to quickly adapt to more complex environments. However, the general curriculum learning  may overfit the current training environments and thus forget  previous skills.

\begin{figure}[t]
\setlength{\abovecaptionskip}{0cm}
\center
\includegraphics[width=0.475\textwidth]{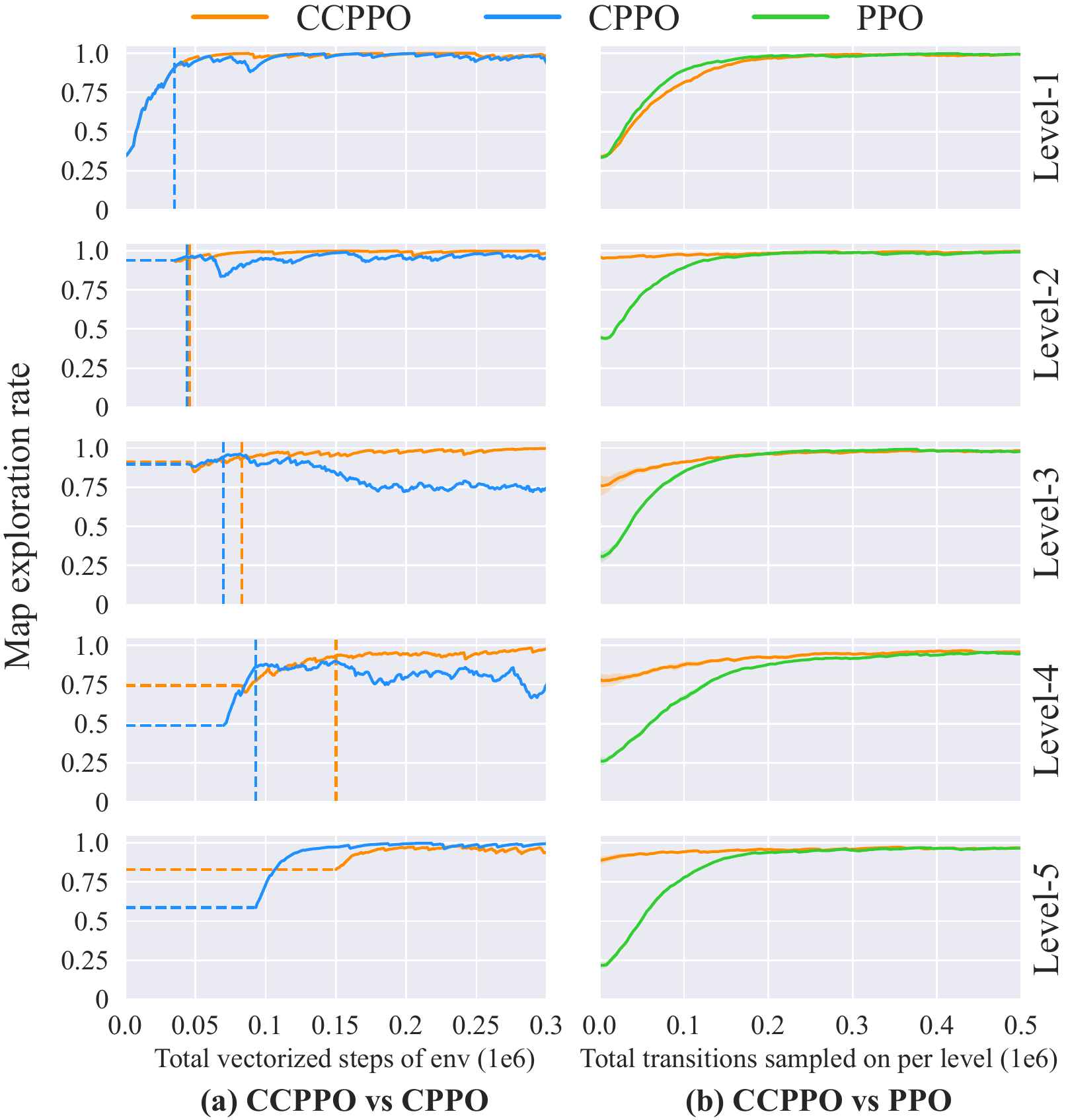}
\caption{\textbf{Training curves in grid-based simulator}: In (a), to highlight the training process of curriculum learning and the issue of catastrophic forgetting, the horizontal coordinate is set to \emph{total vectorized steps of environments}. During the training, the updated policy is evaluated simultaneously on all previously experienced levels with a certain frequency. The vertical dotted line represents the moment of switching to the next training stage, and the horizontal dotted line represents the initial map exploration rate on the current level. Since different random seeds lead to different moments of switching training stages, we only show the training result under one of the five seeds, but other seeds have similar results. In (b), to facilitate the comparison of sample efficiency, the horizontal coordinate is set to \emph{total transitions sampled on per level}. We conduct experiments under five different random seeds, where the solid line represents the average, and the shadow represents the standard deviation. The curves in (a) and (b) are smoothed by the EMA as in equation (5), and $\tau =0.9$.}
\label{fig4}
\vspace{-0.5cm}
\end{figure}

\subsubsection{CCPPO vs PPO}
In this experiment, since PPO does not use curriculum learning, the comparison focuses on sample efficiency, which is defined as the number of transitions sampled on each level when the exponential moving average (EMA) of the map exploration rate first reaches 0.95. The  update rule of EMA is formulated as
\begin{equation}
\bar{\rho}_i=\tau\cdot \bar{\rho}_{i-1} + (1-\tau) \cdot \rho_i ,
\end{equation}
where  $\tau =0.9$, $\rho_i$ is the map exploration rate at the $i$th evaluation and $\bar{\rho}_i$ is the corresponding EMA. We conduct experiments under five different random seeds, the training curves for CCPPO and PPO are shown in Fig. 4(b),  and the sample efficiencies are listed in TABLE \uppercase\expandafter{\romannumeral1}.

\begin{table}[htbp]\scriptsize
\vspace{-0.1cm}
\setlength\tabcolsep{1pt}
\caption{Sample Efficiencies for CCPPO and PPO $(\times 10^3)$}
\label{tab1}
\centering
\begin{tabular}{c c c c c c}
\toprule[1pt]
 & Level-1 & Level-2 & Level-3 & Level-4 & Level-5\\
\midrule[0.5pt]
CCPPO &  170.4 $\pm$ 12.5 & \textbf{13.6} $\pm$ \textbf{16.9} & \textbf{159.2} $\pm$ \textbf{25.0} & \textbf{301.6} $\pm$ \textbf{92.7} & \textbf{68.0} $\pm$ \textbf{62.2}\\
PPO &  \textbf{144.8} $\pm$ \textbf{17.0} & 139.2 $\pm$ 7.8 & 167.2 $\pm$ 7.8 & 413.6 $\pm$ 47.9& 241.6$\pm$ 45.0\\
\bottomrule[1pt]
\vspace{-0.3cm}
\end{tabular}
\end{table}

We can obviously find that the sample efficiencies of CCPPO on the last four levels are all higher than those of PPO. It can be interpreted that CCRL facilitates the accumulation and transfer of knowledge among different tasks to improve sample efficiency, compared with learning from scratch without a curriculum.

\begin{figure}[t]
\setlength{\abovecaptionskip}{0cm}
\center
\includegraphics[width=0.475\textwidth]{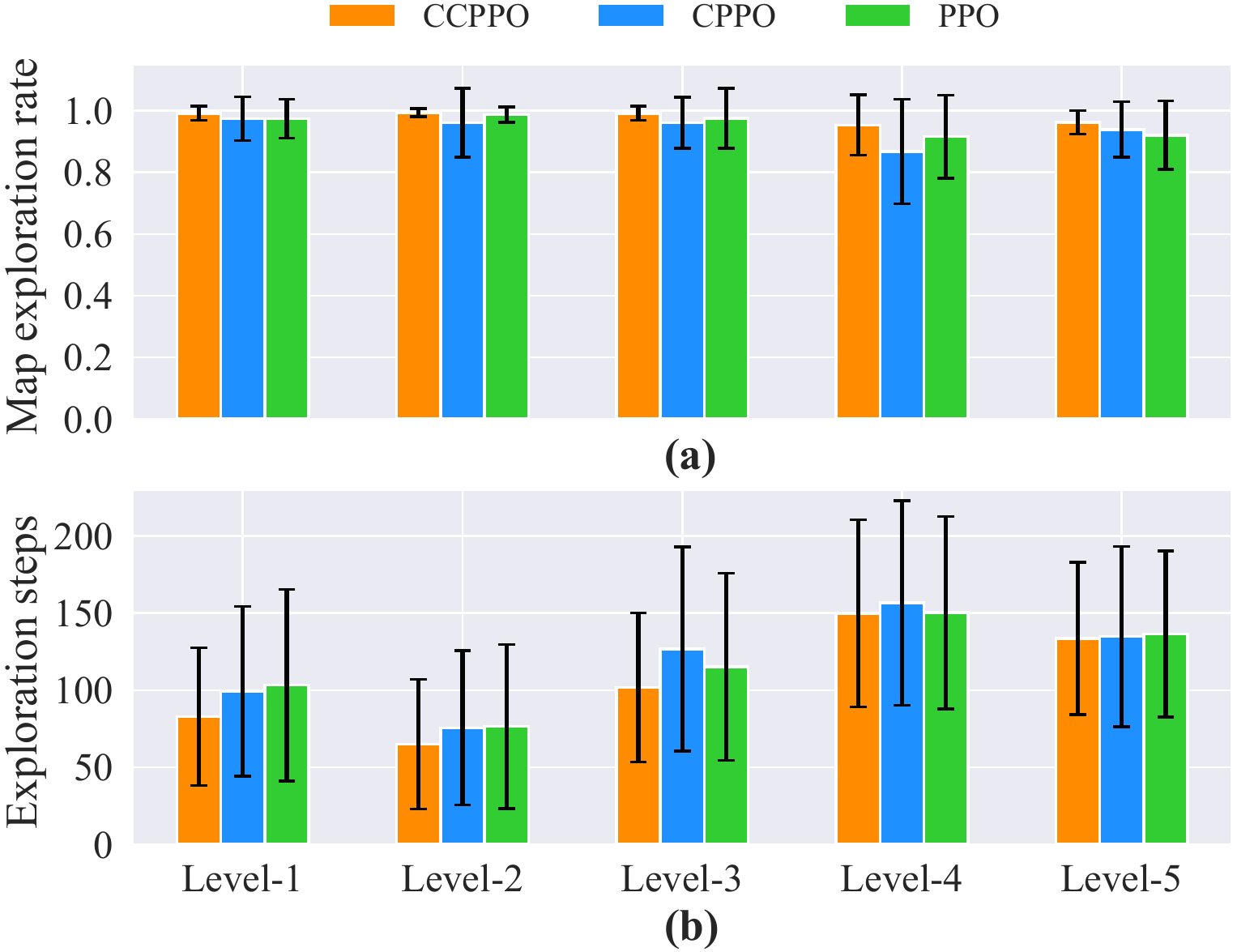}
\caption{\textbf{Statistics of map exploration rates and exploration steps on the test maps:} Histogram means the average value under 100 episodes, and the error bar is the standard deviation. }
\label{fig5}
\vspace{-0.5cm}
\end{figure}

\subsection{Generalization Experiments}
To compare the zero-shot generalization of different algorithms, after training the same number of steps in the training maps, we directly transfer the final models of  CCPPO, CPPO and PPO to a set of test maps with different sizes and layouts (as shown in Fig. 2(b)).  There are two metrics for evaluating the performance of different algorithms:
\begin{itemize}
\item \emph{Map exploration rate}: is defined as the final map exploration rate at the end of an episode, indicating the exploration completeness of the algorithm.
\item \emph{Exploration steps}: is defined as the number of steps when the map exploration rate first reaches 0.95, indicating the exploration efficiency of the algorithm.
\end{itemize}

We run 20 episodes under five random seeds respectively (100 episodes per test map in total), and the statistics of map exploration rates and exploration steps are shown in Fig. 5. It can be found that CCPPO has higher map exploration rates and lower exploration steps on all five test maps compared to CPPO and PPO. The reason is that CCRL focuses more on the accumulation of knowledge during the learning process, making it easier to generalize generic skills to unseen environments, rather than simply memorizing fixed sequences of actions. In addition, CPPO has the worse generalization and exploration efficiency due to the issues of catastrophic forgetting and overfitting in general curriculum learning. The mapping results of CCPPO in the training and test maps are shown in Fig. 6.

\begin{figure}[b]
\vspace{-0.3cm}
\setlength{\abovecaptionskip}{0cm}
\center
\includegraphics[width=0.45\textwidth]{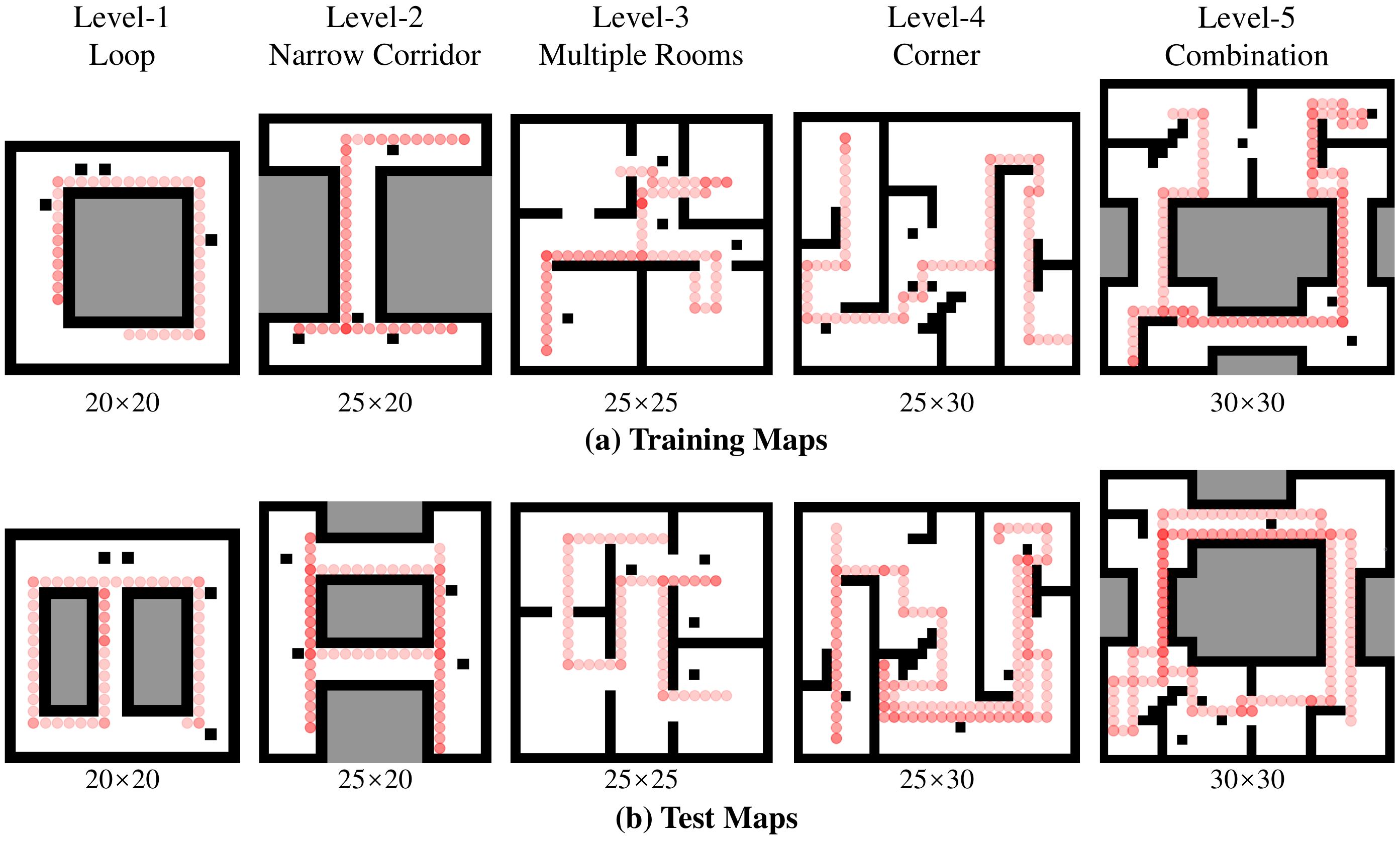}
\caption{\textbf{Mapping results of CCPPO on the training and test maps}. The red circles represent the moving trajectory of the agent during the exploration, and the black boxes are obstacles randomly placed on the maps.}
\label{fig6}
\vspace{-0.1cm}
\end{figure}

\section{CONCLUSIONS}
In this paper, we train an end-to-end autonomous exploration and mapping model based on DRL and curriculum learning. We present an improved state representation that can adapt to different size environments. Besides, we customize a concise grid-based autonomous exploration simulator specifically for end-to-end training and curriculum learning, facilitating fast implementation, verification and comparison of DRL algorithms. In addition, we propose a Cumulative Curriculum Reinforcement Learning (CCRL) training framework to moderate the catastrophic forgetting issue faced by general curriculum learning, which can also improve the sample efficiency and generalization of DRL algorithms, as shown by experimental results. In future research, we will focus on the sim-to-real transfer and conduct experiments in the real world.


\bibliographystyle{IEEEtran}
\bibliography{IEEEexample}

\end{document}